# Sports Camera Pose Refinement Using an Evolution Strategy

Grzegorz Rypeść
*Warsaw University of Technology*
*Sport Algorithmics and Gaming*
Warsaw, Poland
g.rypesc@sagsport.com

Grzegorz Kurzejamski
*Sport Algorithmics and Gaming*
Warsaw, Poland
g.kurzejamski@sagsport.com

Jacek Komorowski
*Warsaw University of Technology*
*Sport Algorithmics and Gaming*
Warsaw, Poland
jacek.komorowski@pw.edu.pl

*Abstract*—This paper presents a robust end-to-end method for sports cameras extrinsic parameters optimization using a novel evolution strategy. First, we developed a neural network architecture for an edge or area-based segmentation of a sports field. Secondly, we implemented the evolution strategy, which purpose is to refine extrinsic camera parameters given a single, segmented sports field image. Experimental comparison with state-of-the-art camera pose refinement methods on real-world data demonstrates the superiority of the proposed algorithm. We also perform an ablation study and propose a way to generalize the method to additionally refine the intrinsic camera matrix.

## I. INTRODUCTION

Automatic analysis of sports games is a broad area covering fields of computer vision, computer graphics, and deep learning [1]. Industry companies offer real-time tracking systems for various sports, e.g. SportVU [2] solution from Stats Perform and Optical Tracking System [3] from ChyronHego TRACAB. These systems rely on the ability to estimate real-world objects' positions from images. Estimation of players' position and speed or detection of events, such as goals or offside, require an ability to map 2D pixel coordinates in the image plane to 3D coordinates in the world reference frame. This can be achieved using a 3x4 projection matrix calculated for each installed camera. If we assume the world model is planar, i.e., the sports field is a flat surface, then a 3x3 homography matrix is enough to perform the mapping. In each case, these matrices can be calculated using intrinsic and extrinsic camera parameters. Intrinsic parameters, specify internal cameras parameters, such as a camera focal length and a principal point coordinates. They are represented as a 3x3 matrix, defining the transformation from 3D point coordinates in the camera reference frame, to the pixel coordinates in the image plane. Extrinsic parameters, referred to as a camera pose, describe the camera rotation and translation with respect to the world coordinate frame. They define the transformation from a 3D point coordinates in the world coordinate frame to 3D coordinates in the camera reference frame.

Sports cameras are initially calibrated after they are installed around the playing field. Their intrinsic and extrinsic parameters are estimated by performing a manual calibration procedure using a planar chessboard-like calibration pattern and Zhang method [4]. However, in an outdoor environment cameras are affected by an adverse environmental conditions. A strong wind or temperature variations cause small changes to the camera position and viewing angle. Projection and homography matrices estimated during an initial camera calibration become inaccurate over time, which leads to an incorrect conversion from camera coordinates to world coordinates, as shown in the left part of Fig. 1. This adversely affects the quality of the entire sport analytic pipeline. A manual calibration process is relatively difficult and very time consuming, and it's not feasible to perform it frequently to correct camera calibration errors. To solve this problem, we present a method for automatic refinement of cameras extrinsic parameters using an evolutionary algorithm. This allows a quick correction of camera extrinsic parameters before each event. Intrinsic camera parameters (e.g. focal length) are less susceptible to environmental conditions and remain relatively constant.

Our method consists of two stages: image segmentation and extrinsic parameters refinement. Firstly, we segment the sports field image to find robust field marking, such as sports field lines or areas, using a deep convolutional neural network. We call these markings robust as players do not occlude them in a significant way because of their size. The idea for the second stage is to iteratively warp segmented image into bird's-eye view using camera parameters and evaluate how well it's aligned with the field template. If the matching is not perfect, then we can refine extrinsic parameters to make the alignment better. To model this iterative process mathematically, we define the problem as an optimization task in a 6-dimensional space, with 3 translation and 3 rotation degrees of freedom. The objective is to minimize the previously defined intuitive loss function. Inspired by biology, we propose performing the optimization using an evolution strategy. Section IV shows that this proves to be a better solution than gradient descent-based approaches. It is important to note that our method requires a reasonably well initialization of camera extrinsic parameters. For example, the starting parameters cannot define the camera looking in the opposite direction compared to the ground truth. In this work, we investigate the robustness of our method to the extrinsic parameters initialization.

To the best of our knowledge, evolution strategies involving sports field images segmented with deep learning have never been used for extrinsic camera calibration. We demonstrate that this is an efficient and effective solution, performing better than gradient descent-based algorithms in real-life scenarios. Experimental evaluation, using our real-world dataset with football pitch images, proves the superiority of our method compared to competitive approaches. To summarize, our contributions are:

- proposal of edge and area-based deep convolutional architectures for sports field edge and area segmentation,
- definition of evolution strategy and fitness function for a robust camera extrinsic calibration.

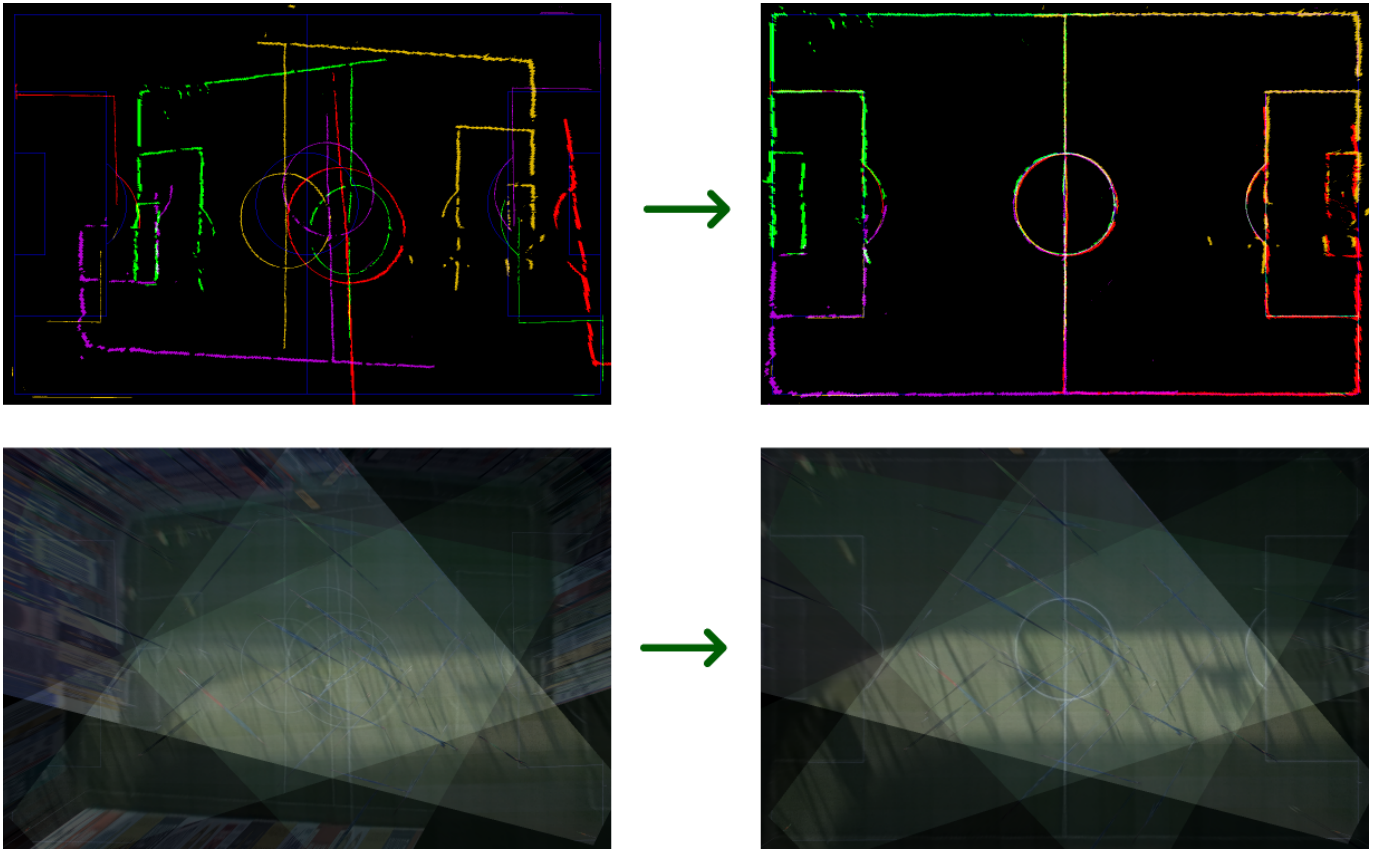

Fig. 1. Camera pose refinement results of a system of four outdoor cameras. Cameras were installed in a soccer field corners and their parameters were not updated for three months. Left side shows soccer field images (bottom) and segmented field lines (top) warped to a bird's-eye view using initial extrinsic camera parameters. Right side shows images warped to a bird's-eye view after performing pose refinement using our method.

The structure of this paper is as follows. We discuss related work in section 2, in section 3 we describe our method, and evaluate it on a soccer dataset in section 4. We also compare it with state of the art. Finally, we summarize the paper and conclude further works in section 5.

## II. RELATED WORK

### A. Sports Camera Calibration

The topic of sports camera calibration has been widely discussed in the literature. Most of the approaches focus on the problem of estimating the homography matrix or relocalizing the camera, i.e. finding its pose. Early approaches to these problems [5]–[8] use Hough transforms [9] to extract geometric primitives such as lines, arcs and circles from images. These primitives allow finding control points or use a combinatorial search to estimate camera parameters. Unfortunately, these methods require a lot of manual work in order to parametrize color and texture-based heuristics for geometric primitives detection. These problems are resolved by a progress in deep learning. Deep neural convolutional networks, such as U-Net [10], allow automatic semantic segmentation of images, without the need for constructing hand-crafted features. Sha et. al. [11] uses such network to perform the area-based court segmentation. Image segmentation allows for an initial pose estimation using a Siamese network to calculate a corresponding entry in a feature-pose dictionary/database. Such database contains pairs of edge images showing segmented field lines with corresponding camera poses. Finally, a homography is refined using spatial transformer network [12]. A similar approach is described in [13], where generative adversarial networks [14] are used to detect sports field markings. Camera pose is initially estimated using a feature-pose database and refined using Lucas-Kanade algorithm and chain rule calculated on distance images. The drawback of these approaches is an assumption that refining transformations are small and local.

An iterative approach to calibration is described in [15].

Firstly, a deep convolutional network is trained, which takes a sports field template and a camera image as an input. Its output is an estimation of calibration error, e.g. IoU between sports field template and a ground truth template in the camera view. Then, the calibration process uses a bilinear sampler to create a fully differential computing graph from an estimated homography matrix to the calibration error. Stochastic gradient descent algorithm allows to backpropagate the approximated calibration error and iteratively updates the homography to lower it. Such approach is prone to be stuck in a local minimum as search space is not convex.

An approach to sports camera calibration using RANSAC [16] and the genetic algorithm is described in [17]. Field lines are detected in an input image and their intersections define control points used by the fitness function in the genetic algorithm. The result of the optimization process defines new camera extrinsic and intrinsic parameters.

*B. Evolutionary algorithms for image registration*

Image registration is a similar task to camera relocalization, where the problem is to estimate a transformation of a set of different images to the same coordinate system. In this area, there exist known evolutionary approaches. In [18] the problem of satellite image registration is modeled as an optimization problem in the search space of affine transformation parameters. A differential evolution algorithm is used to solve it. Ma et. al. [19] proposes a similar mathematical model for the remote sensing image registration but uses an orthogonal learning differential evolution algorithm as an optimizer.

### III. METHOD

*A. Problem definition*

We define the extrinsic camera calibration process as a heuristic search in $\mathbb{R}^6$ space. First three dimensions define the rotation using a Rodrigues representation, $\vec{r} \in \mathbb{R}^3$. Last three dimensions define a translation vector $\vec{t} \in \mathbb{R}^3$. Therefore, elements of the search space are $[r_0, r_1, r_2, t_0, t_1, t_2]$ vectors. We assume that the camera's intrinsic parameters are correct and do not change during the calibration process. They are estimated earlier, during an initial camera calibration, using e.g. a planar chessboard-like calibration pattern and the Zhang [4] method.

Additionally, we model camera extrinsic noise as a multivariate random vector $\vec{\xi} \in \mathcal{N}(0, \xi_r)^3 \times \mathcal{N}(0, \xi_t)^3$ which was undesirably added to valid calibration parameters and caused calibration errors. $\xi_r$ denotes the standard deviation for the rotation vector and $\xi_t$ is the standard deviation for the translation vector. Such noise can be a caused by camera undesired movement due to a strong wind or thermal expansion of camera materials. Our study shows that for football sports cameras, $\xi_t$ is approximately 20 times bigger than $\xi_r$.

High-level view of our pose refinement method is shown in Fig. 2. It consists of two stages: semantic segmentation of a play field image and evolutionary optimization of camera extrinsic parameters. In the first stage, a deep neural network is used to segment an input image. The segmented image is blurred and passed to the second stage, which is an iterative optimization process using evolution strategy. Alternatively, in the second stage we can use other optimization techniques. In section IV we compare the evolution strategy with a stochastic gradient descent (SGD) and Adam [20] optimization algorithms. In our calibration pipeline, we assume a known planar sports field template and known estimates of $\vec{\xi_r}$ and $\vec{\xi_t}$, which can be obtained from historical data or adjusted empirically. We also assume, that we are given starting extrinsic camera parameters in form of a vector $[\vec{r} + \vec{\xi_r}, \vec{t} + \vec{\xi_t}]$.

*B. Homography calculation from camera parameters*

Homography matrix $H$ in computer graphics is a 3x3 matrix that allows performing a homography, i.e an isomorphism that transforms a point from one planar surface to another. The isomorphism can be reversed using $H^{-1}$. Each camera's homography matrix allows performing a projection from a camera plane onto the sports field plane in a bird's-eye view. According to the camera pinhole model, it is possible to calculate H from $\vec{r}$, $\vec{t}$, and camera intrinsic matrix $K$.

Firstly, a rotation vector $\vec{r}$ must be converted into 3x3 rotation matrix $R$ that has 3 degrees of freedom using a Rodrigues method [21]. Secondly, 3x4 projection matrix $P$ is calculated as follows:

$$P = \begin{bmatrix} K_{0,0} & 0 & K_{0,2} \\ 0 & K_{1,1} & K_{1,2} \\ 0 & 0 & 1 \end{bmatrix} \begin{bmatrix} R_{0,0} & R_{0,1} & R_{0,2} & t_0 \\ R_{1,0} & R_{1,1} & R_{1,2} & t_1 \\ R_{2,0} & R_{2,1} & R_{2,2} & t_2 \end{bmatrix} \quad (1)$$

Because the assumption is that sports field is planar, third column of P, which corresponds to the z axis (perpendicular to the plane) can be dropped and the result can be normalized, thus giving us homography matrix $H$:

$$H = \begin{bmatrix} P_{0,0} & P_{0,1} & P_{0,3} \\ P_{1,0} & P_{1,1} & P_{1,3} \\ P_{2,0} & P_{2,1} & P_{2,3} \end{bmatrix} / P_{2,3} \quad (2)$$

*C. Semantic segmentation*

The purpose of the semantic segmentation model, denoted by function $\Psi$, is to segment an input camera image $I$ into an image $\Psi(I)$ of the same size containing only relevant sports field features. Segmented features can be sports field lines or areas such as a penalty area. A segmented image is needed for the second stage of the calibration process, where it is used to calculate the value of the fitness function by measuring alignment with a blurred sports field template $T$. To obtain $T$, we create multiple images from the sports field template, blur each one by a Gaussian function of a different radius and sum them all together. Such transformation makes the fitness function smoother and search space easier to explore for our algorithm. We prove this intuition to be right in our ablation study.

For $\Psi$, we use a modified U-Net architecture [10] with sigmoid activation function after output layer and additional padding of 1 in encoder layers to ensure that $\Psi(I)$ is of the same size as $I$. Maintaining the image size is crucial for calibration quality. We also use PReLu instead of ReLu

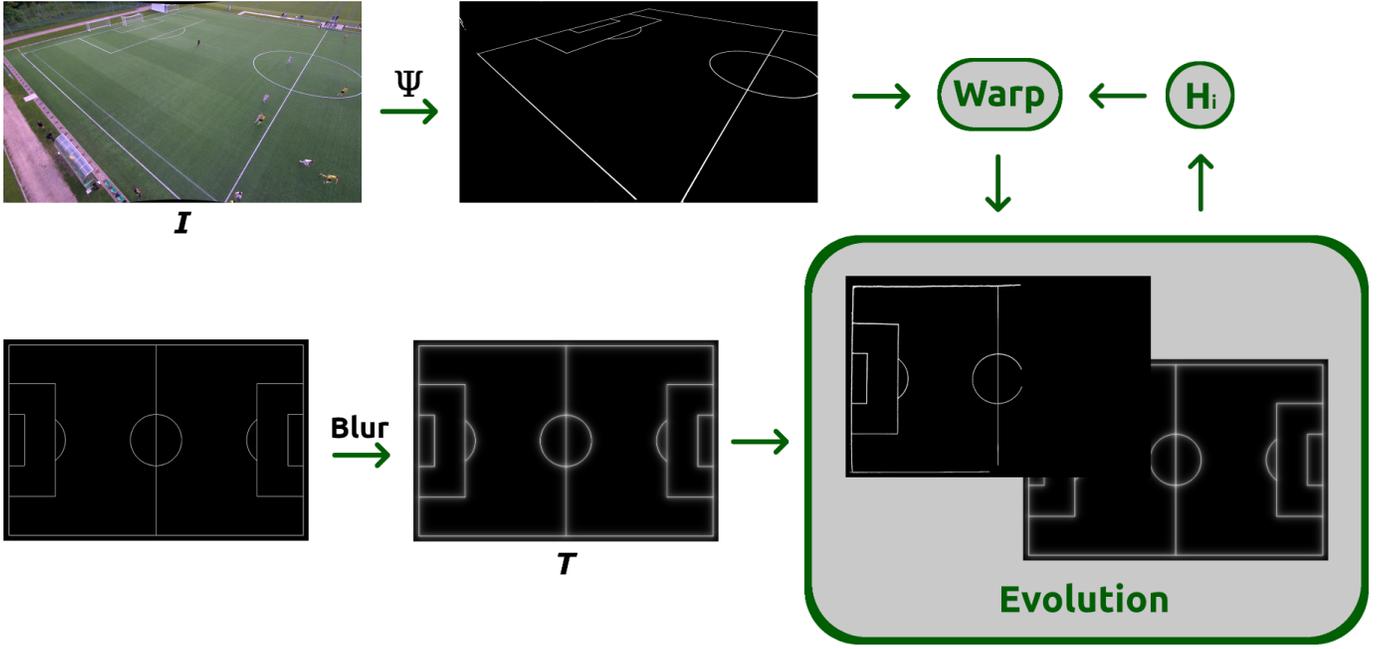

Fig. 2. Overview of our camera pose refinement method. Firstly, the camera image $I$ is segmented using the deep convolutional network, denoted by the function $\Psi$. The sports field template is blurred using Gaussian functions of different radii; the clipped sum of these transformations produces $T$. Then, in each generation, the evolution strategy warps segmented images for each individual $i$ with a homography matrix $H_i$ calculated from one's rotation, translation vectors, and camera intrinsic matrix. This allows calculating fitness function between $H_i(\Psi(I))$ and $T$ in top view perspective and selecting individuals that provide better calibration. They move to the next generation, or if it is the last generation, the best one defines new camera extrinsic parameters and optimization process stops.

activation functions. In our study, PReLu allowed achieving better results.

The evolution strategy we present can work with different types of segmented images. We train and compare two independent models. First segments only field lines. It classifies each pixel into one of two classes: line and not line. The second model performs football field area segmentation. It classifies each pixel into one of 4 classes: belonging to a penalty area; a goal area; other area belonging to the sports field; and the area outside the sports field. In the case of the semantic segmentation of football field lines, there is a strong imbalance between the foreground (field lines) and background pixels. What's more, lines that are further away from the camera appear smaller in images and contribute less to standard loss functions than closer lines. To compensate for this problem we use $\alpha$-balanced variant of a focal loss [22] as the objective function during training:

$$L_{FL} = -\alpha(1-p_c)^\gamma \log(p_c) \quad (3)$$

where $\alpha$ and $\gamma$ are adjustable hyperparameters. Given $p$ as a probability of a pixel belonging to a class with label $y = 1$ (binary classification), $p_c$ is defined in the following way:

$$p_c = \begin{cases} p & \text{if } y = 1 \\ 1 - p & \text{otherwise} \end{cases} \quad (4)$$

### D. Evolutionary optimization

Our evolution strategy is based on $(\mu + \lambda)$ elitist evolution strategy, where in each generation $g \in 0, 1, ..., G-1$ all $\mu$ parents, denoted as $P_g$ are randomly recombined to create $O_g$ offspring of size $\lambda$. The optimization is done for G generations. The starting population with $\mu$ individuals is created by taking an initial estimate of camera extrinsic parameters $[\vec{r}+\vec{\xi_r}, \vec{t}+\vec{\xi_t}]$ and randomly mutating it by corrupting with a Gaussian noise $\vec{\xi}$. In each generation, an offspring is created consisting of $\lambda$ children. Due to recombination, each child takes form of a vector $[\vec{a} * \vec{p_i} + (\vec{1} - \vec{a}) * \vec{p_j}]$ where $a \in \mathcal{U}(0,1)^6$, $p_i$ and $p_j$ are randomly sampled parent vectors from $P_g$. The offspring is further mutated using random vector $\mathcal{N}(0, \xi_{decay}^g \xi_r)^3 \times \mathcal{N}(0, \xi_{decay}^g \xi_t)^3$ where $\xi_{decay}$ is a decay factor of mutation strength. This factor allows to improve exploration of the search space in first generations and later increase exploitation of found minima. After the mutation step, a homography matrix $H_i$ is calculated for each offspring individual according to Eq. (2). Then, the whole population of $\mu + \lambda$ individuals is sorted in ascending order according to the fitness function. The best $\mu$ individuals move on to the next generation, and the rest is terminated, what guarantees that the adaptation quality in each generation is at least as good as in the previous generation. Finally, after the last generation, the new extrinsic parameters of the camera are set to the parameters of the best individual in $P_{G-1}$. We provide pseudocode for the algorithm in Fig. 3.

The optimization goal of the refinement process is to find a minimum for the fitness function defined as follows:

$$Fitness = \frac{\Sigma[H(\Psi(I)) - T * B]^{\circ 2}}{\Sigma(B)} \quad (5)$$

1. Initialize $T, \mu, \lambda, \xi_r, \xi_t, \xi_{decay}, G, K, P_0 = \{i_0..., i_{\mu-1}\}$
2. Calculate $H_i$ for each individual in $P_0$
3. Calculate the fitness function for each individual in $P_0$
**for** $g \in 0, ..., G-1$ **do**
   4. Create offspring $O_g$ by randomly sampling 2 parents from $P_g$ for each child and interpolating them
   5. Mutate $O_g$ with $\mathcal{N}(0, \xi_{decay}^g \xi_r)^3 \times \mathcal{N}(0, \xi_{decay}^g \xi_t)^3$
   6. Calculate $H_i$ for each individual in $O_g$
   7. Calculate the fitness function for each individual in $O_g$
   8. $P_g = P_g \cup O_g$
   9. Sort $P_g$ according to fitness in an ascending order
   10. Leave in $P_g$ only top $\mu$ individuals
**end for**
11. Set camera extrinsic parameters as $i_0$

Fig. 3. Evolution strategy for camera pose refinement pseudocode.

where $\Sigma$ is a function that sum over all pixels, $H(\Psi(I))$ is a segmented image warped into bird's-eye view, $*$ denotes element-wise multiplication, $\circ 2$ stands for Hadamard square power and B is a binary mask in which each element is 1 if corresponding element in $H(\Psi(I))$ is above 0.5. Otherwise, the element is 0. This fitness function produces scalar values, which can be intuitively interpreted as a level of alignment of a segmented image warped into a bird's-eye view with the blurred template image. In our study, this function performed more stable than mean squared error (MSE) between $H(\Psi(I))$ and $T$.

## IV. EVALUATION AND EXPERIMENTS

### A. Dataset

We use an in-house dataset gathered using a multiple-camera image acquisition set installed at football academies. The dataset consists of 1020 RGB images in 4K resolution (3840x2160 pixels) from real-world football games. Images come from 34 videos recorded with different cameras set on six different football fields. We divided the set of videos into train, validation, and test subsets in the proportion of 28:3:3, respectively. 30 images are randomly sampled from each of the recordings. There are also 34 sets of ground truth camera parameters provided, i.e. camera translation, rotation, and intrinsic matrices. They were calculated using a manual selection of control points and conventional algorithms: Levenberg-Marquardt iterative optimization and Zhang method. To simulate calibration errors, we perturb intrinsic parameters with $\vec{\xi}$ defined in Section III. We test calibration algorithms twice for different distortion rates $\xi_r$ and $\xi_t$: (0.005, 0.1) and (0.015, 0.3). Calibration errors that are created due to the perturbations are presented in Table III as $Start$.

### B. Metrics

We use well-known metrics such as intersection over union (IoU), recall, precision, F1-score defined for the binary classification task to evaluate semantic segmentation models.

To evaluate the results of the pose refinement algorithms, we use two metrics:

*a) $IoU_{part}$:* A metric that is popular in literature [11], [15], [23] and is computed in a camera view. To calculate $IoU_{part}$, we calculate intersection over union between 2 sports fields templates. First is warped into camera view using ground truth homography. The second is warped according to the estimated homography.

*b) Real world error score set (RWE):* An important metric in player and ball tracking problems is an error measured in the real-world metric, such as centimeters. For tasks that require high accuracy, such as player tracking, this error should be as low as possible to ensure calibration error will not propagate into the player tracking pipeline. Therefore, we define a $RWE$ set that consists of real-world error distances between ground truth points and the same points warped into camera view using ground truth homography $H_{gt}$ and then warped back to the world model using an inverse of estimated homography $H_{est}$. The set allows the calculation of real-world calibration error statistics. The ground truth points are sports field keypoints visible in camera view, denoted as $KP$. These are center, corners, and halfway line intersections with sidelines. We assume these points belong to the same plane. Given a set of images and euclidean distance function $d$, the $RWE$ set is defined as follows:

$$RWE = \{d(p, H_{est}^{-1} H_{gt} p) : p \in KP \land H_{gt}(p) \in I\} \quad (6)$$

To evaluate calibration algorithms, we report the mean and standard deviation of the $RWE$ set.

### C. Baselines

We compare our method to the refinement part of the pipeline described in [15] denoted by OTLE. This approach makes use of a regressor that estimates $IoU_{part}$ between sports field visible in image and sports field template warped into the camera's plane of view. It uses the gradient descent method to refine rotation and translation vectors to increase the $IoU_{part}$. The authors did not make the training code public, but they shared the implementation of their ResNet-18 [24] model with spectral normalization. We trained this network from scratch using Adam optimizer and a batch size of 4. We denote this network as ResNet-18+SN. For its training, we initially set the learning rate to 0.001 and decrease it by 2% every epoch. We chose a model that had the lowest validation MSE error from 100 epochs of training. For each training dataset, we apply noise of $\vec{\xi}$ and data augmentation. During inference, we use Adam optimization algorithm with default hyperparameters and a learning rate of 1e-4 to update extrinsic parameters to lower the estimated calibration error. We utilize the grid sampler with bilinear interpolation to maintain the differentiability of the estimated $IoU_{part}$. We run Adam for 50 iterations.

OTLE described earlier operates on an RGB image and a template. Our method produces mid-way byproducts in the form of binarized line masks with U-Net architecture, which may benefit the OTLE-like approach. Thus, we also perform an experiment where instead of using the evolution strategy to lower the fitness function (5) we use the SGD algorithm and optimize $\vec{r}$ and $\vec{t}$ vectors in a similar manner to OTLE. We are not using an evolution strategy at all for this baseline. It is run for 50 epochs with a learning rate of 0.0001 and momentum of 0.9. We denote these methods as U-Net+SGD and U-Net-Area+SGD.

### D. Our method hyperparameters

We denote our segmentation models as U-Net and U-Net-Area. The former performs sports field lines segmentation and the latter sports field area segmentation. These models have four times fewer filters in each layer than the original U-Net architecture. They are trained for 100 epochs with the same optimizer and learning rate schedule described for the ResNet-18 $IoU_{part}$ regressor. We denote evolution strategy utilizing outputs from line segmentation as U-Net+ES and outputs from area segmentation as U-Net-Area+ES. For both of them we set $\mu$ to 64, $\lambda$ to 128, G to 50, $\xi_{decay}$ to 0.95, $\xi_r$ and $\xi_t$ to those set for dataset.

### E. Results

Results are divided into 3 categories: results of semantic segmentation models (Table I), $IoU_{part}$ regressors (Table II) and pose refinement results (Table III). All tests were performed using PC with AMD EPYC[TM] 7401P and Nvidia GeForce RTX[TM] 3090. An average time of pose refinement for our U-Net+ES method is approximately 14 seconds.

TABLE I
RESULTS OF SEGMENTATION MODELS

|  | IoU | Recall | Precision | F1 |
|---|---|---|---|---|
| U-Net | 0.29 | 0.51 | 0.35 | 0.41 |
| U-Net-Area | 0.78 | 0.86 | 0.88 | 0.87 |

TABLE II
RESULTS OF IoU REGRESSION MODELS

|  |  | MSE | MAE |
|---|---|---|---|
| $\xi_r = 0.005$ | ResNet-18 | $2.93 \cdot 10^{-4}$ | 0.0171 |
| $\xi_t = 0.1$ | ResNet-18 + SN | $3.65 \cdot 10^{-4}$ | 0.0191 |
| $\xi_r = 0.015$ | ResNet-18 | $2.21 \cdot 10^{-3}$ | 0.0470 |
| $\xi_t = 0.3$ | ResNet-18 + SN | $4.64 \cdot 10^{-3}$ | 0.0681 |

Line-based segmentation model (U-Net) performed significantly worse compared to an area-based segmentation model (U-Net-Area). This can be attributed to imperfections in the training dataset. As playing field lines are very thin, even small errors in the ground truth annotations cause mislabelling of a relatively large number of pixels belonging to a field line in the image.

However, using our evolutionary pose refinement method with line-based segmentation (U-Net+ES) yielded the best results. It achieved superior results compared to other, non-evolutionary, methods. New extrinsic camera calibrations it generated were near-perfect, proving robustness of the proposed method.

OTLE+SN, U-Net-Area+SGD and U-Net+SGD based methods in each of 90 test runs fully converged each time but to the wrong local minima. Once stuck there, the loss would no longer decrease, thus making the algorithms fail to recalibrate cameras properly. For small extrinsic camera perturbations ($\xi_r = 0.005$ and $\xi_t = 0.1$), U-Net-Area+SGD performed the best out of all gradient descents methods. It got worse results in the second test ($\xi_r = 0.015$ and $\xi_t = 0.3$) where OTLE+SN performed better.

TABLE III
RESULTS OF CALIBRATION ALGORITHMS

|  |  | $IoU_{part}$ | RWE mean | RWE std |
|---|---|---|---|---|
| $\xi_r = 0.005$ $\xi_t = 0.1$ | Start | 97.62 | 2.30 | 4.07 |
|  | U-Net+ES (ours) | **99.25** | **0.17** | **0.11** |
|  | U-Net-Area+ES (ours) | 98.78 | 0.57 | 0.58 |
|  | U-Net+SGD | 97.62 | 2.44 | 4.46 |
|  | U-Net-Area+SGD | 98.23 | 1.29 | 2.12 |
|  | OTLE | 97.61 | 2.27 | 4.16 |
|  | OTLE+SN | 97.50 | 2.41 | 4.17 |
| $\xi_r = 0.015$ $\xi_t = 0.3$ | Start | 93.15 | 8.05 | 21.90 |
|  | U-Net+ES (ours) | **99.02** | **0.20** | **0.13** |
|  | U-Net-Area+ES (ours) | 98.20 | 0.68 | 0.78 |
|  | U-Net+SGD | 93.19 | 8.11 | 22.08 |
|  | U-Net-Area+SGD | 93.86 | 7.19 | 23.98 |
|  | OTLE | 93.28 | 5.25 | 6.72 |
|  | OTLE+SN | 93.62 | 5.57 | 7.17 |

### F. Ablation study

In our ablation study, we tried to improve the unsatisfactory performance of the OTLE+SN algorithm by using our implementation of Resnet-18 without spectral normalization. We trained it using the same hyperparameters and denote this method as OTLE. This model achieved a much lower regression error, but the improvement in pose refinement task was not significant, 5% lower mean of $RWE$ set.

We also investigated how blurring of the sports field template impacts results of ours U-Net+ES approach. We tested the algorithm on the test set for different amount of Gaussian kernels (0, 1, 3, 7, 15) and different sizes of the kernels. Amount of Gaussian filters equal to 0 means no blur was performed, 1 means that a single kernel was applied to the template. For number of kernels 3, 7 and 15 the template blurring yielded the respective number of images, each transformed with the kernel of different size. First image has no blur (size of kernel is 0), second kernel has the base kernel radius size which is a hyperparameter. The rest of kernels have size of one plus a multiple of base kernel radius size decreased by one. E.g. number of kernels 7 and base kernel size 5 means, that the template was blurred 7 times with kernels of size (0, 5, 9, 13, 17, 21, 25), each time yielding a different image. These images were summed up and clipped to the maximum pixel value, thus giving us the blurred template $T$. For this ablation study we set $\xi_r = 0.015$, $\xi_t = 0.3$, $\mu$=32, $\lambda$=64, $G$=60.

Results are presented in Fig. 4. They prove our intuition concerning that blur positively impacts the search space of the problem. With no blur $RWE$ set has a big standard deviation, therefore the calibration results are unstable. A single kernel is also not enough since median of $RWE$ set and of $IoU_{part}$ is noticeably worse than for larger amount of kernels. For number of kernels 3, 7, 15 and different kernel sizes we observed solid and stable results.

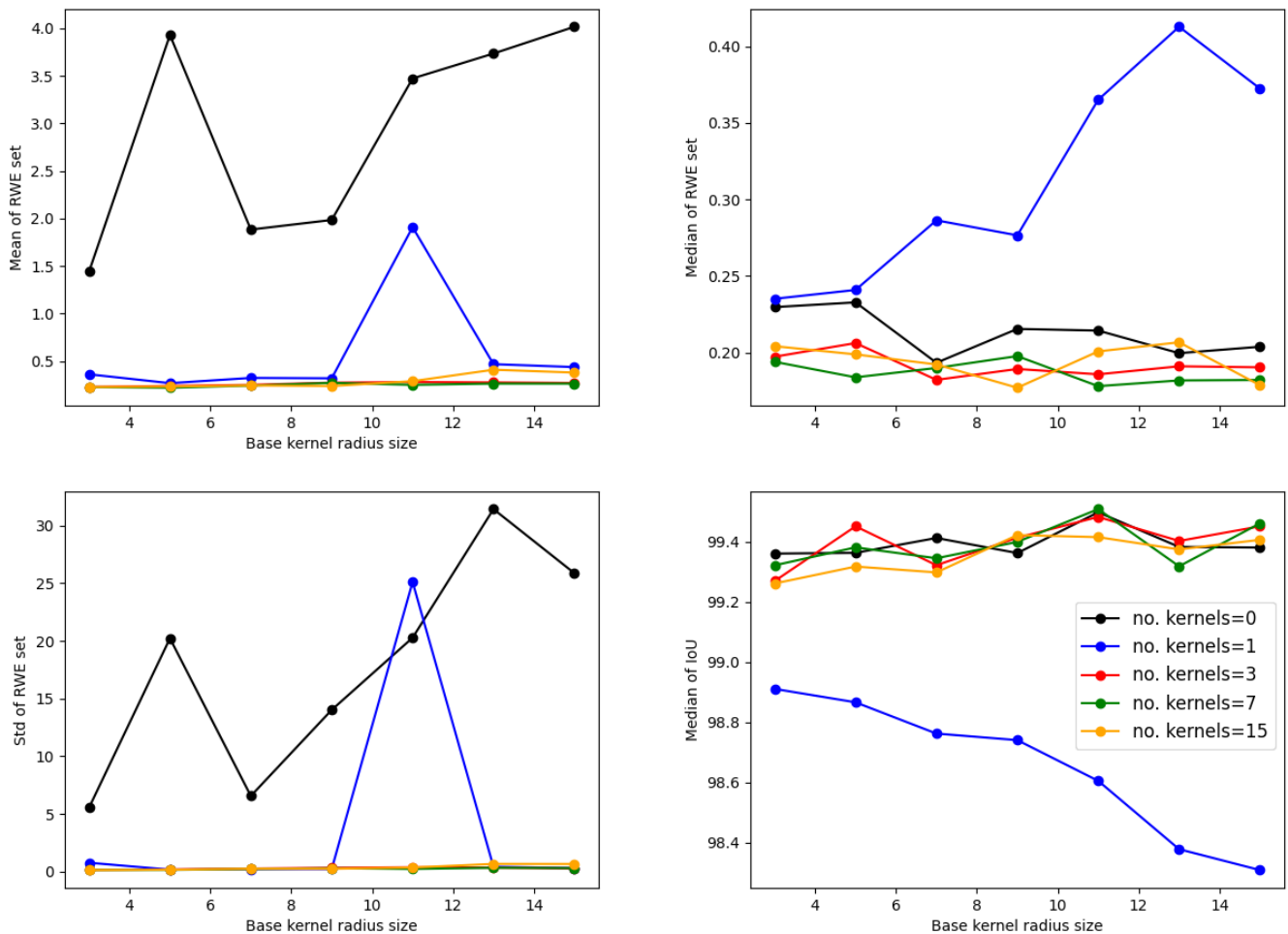

Fig. 4. Mean, median and standard deviation of $RWE$ set and median of $IoU_{part}$ with respect to the number of Gaussian kernels and their size.

From theoretical point of view, the fitness function (5) is a step function. Blurring the template increases function's number of intervals, thus making the search space easier to explore. With blur, even a small step in the space results in a change of value of the fitness function.

## V. Conclusion

In this paper, we propose the robust and accurate solution for the static sports cameras pose refinement problem. The use of evolution strategy avoids being stuck in local minima of the extrinsic parameters search space, thus performing better than gradient descent approaches.

We would like to notice that our camera pose refinement method can be generalized to adapt intrinsic camera parameters by adding them to the definition of the individual. The method can also be changed to operate only on a homography matrix, thus making an individual be a 3x3 matrix with 8 degrees of freedom. Our approach can also be adapted to pose refinement of a set of sports cameras, which view is stitched. In order to achieve the best stitch quality, the fitness function should include additional loss for the part of the sports field that is visible in each camera.

## VI. Acknowledgments

This study was prepared within realization of the Project co-funded by polish National Center of Research and Development, Ścieżka dla Mazowsza/2019.

## References

[1] G. Thomas, R. Gade, T. B. Moeslund, P. Carr, and A. Hilton, "Computer vision for sports: Current applications and research topics," *Computer Vision and Image Understanding*, vol. 159, pp. 3–18, 2017, computer Vision in Sports. [Online]. Available: https://www.sciencedirect.com/science/article/pii/S1077314217300711
[2] Stats Perform, "SportVU 2.0 Real-Time Optical Tracking," 2021. [Online]. Available: https://www.statsperform.com/team-performance/football-performance/optical-tracking
[3] ChyronHego TRACAB Technologies, 2021. [Online]. Available: https://tracab.com/products/tracab-technologies
[4] Z. Zhang, "A flexible new technique for camera calibration," *IEEE Transactions on pattern analysis and machine intelligence*, vol. 22, no. 11, pp. 1330–1334, 2000.
[5] H. Kim and K. S. Hong, "Soccer video mosaicing using self-calibration and line tracking," in *Proceedings 15th International Conference on Pattern Recognition. ICPR-2000*, vol. 1. IEEE, 2000, pp. 592–595.

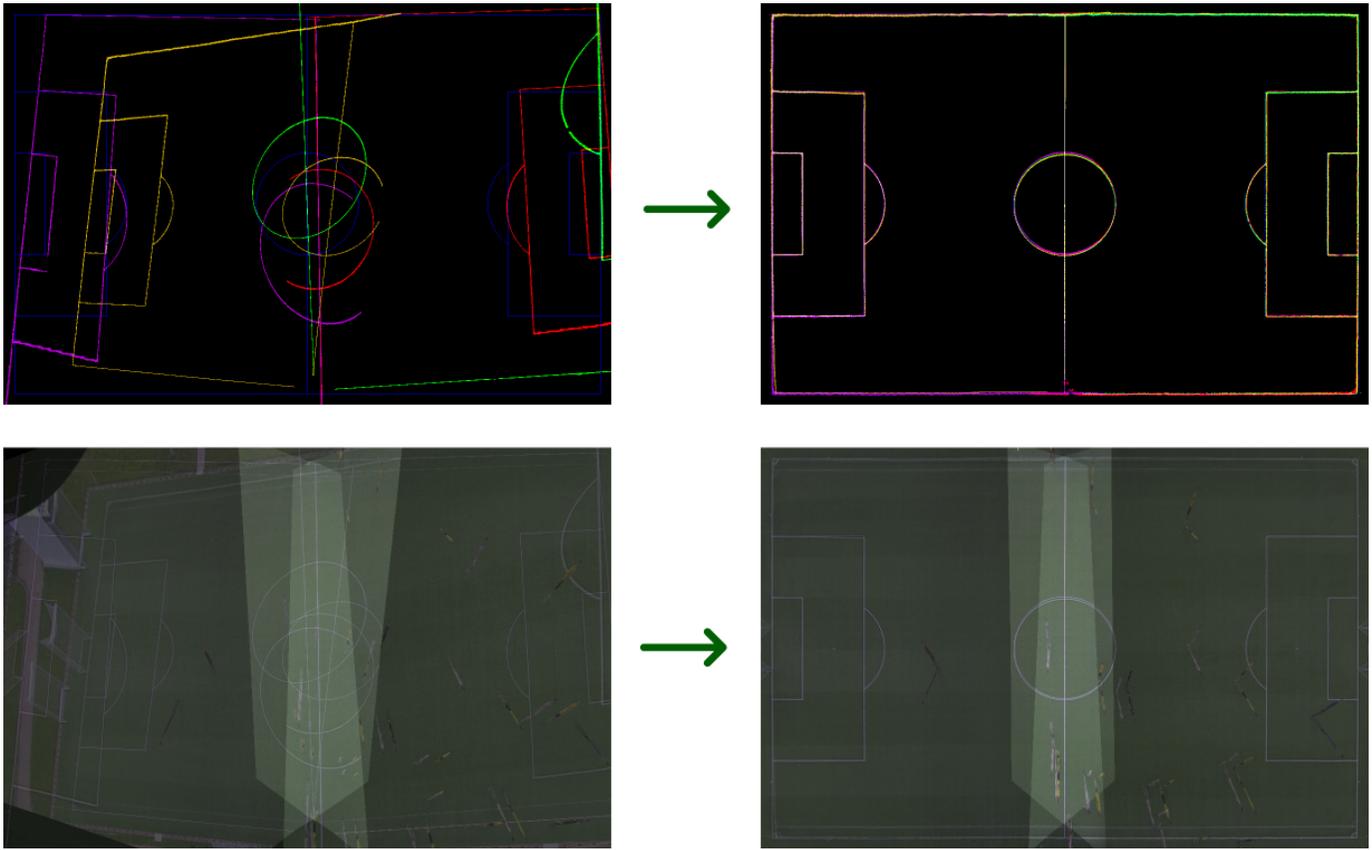

Fig. 5. Results of calibration of a system of 4 cameras using our method. These cameras are placed behind intersections of the middle line and sidelines. Their extrinsic parameters were perturbed with random vector $\vec{\xi}$ of $\xi_r = 0.015$ and $\xi_t = 0.3$.

[6] A. Yamada, Y. Shirai, and J. Miura, "Tracking players and a ball in video image sequence and estimating camera parameters for 3d interpretation of soccer games," in *Object recognition supported by user interaction for service robots*, vol. 1. IEEE, 2002, pp. 303–306.

[7] D. Farin, S. Krabbe, W. Effelsberg *et al.*, "Robust camera calibration for sport videos using court models," in *Storage and Retrieval Methods and Applications for Multimedia 2004*, vol. 5307. International Society for Optics and Photonics, 2003, pp. 80–91.

[8] F. Wang, L. Sun, B. Yang, and S. Yang, "Fast arc detection algorithm for play field registration in soccer video mining," in *2006 IEEE International Conference on Systems, Man and Cybernetics*, vol. 6. IEEE, 2006, pp. 4932–4936.

[9] R. O. Duda and P. E. Hart, "Use of the hough transformation to detect lines and curves in pictures," *Communications of the ACM*, vol. 15, no. 1, pp. 11–15, 1972.

[10] O. Ronneberger, P. Fischer, and T. Brox, "U-net: Convolutional networks for biomedical image segmentation," in *International Conference on Medical image computing and computer-assisted intervention*. Springer, 2015, pp. 234–241.

[11] L. Sha, J. Hobbs, P. Felsen, X. Wei, P. Lucey, and S. Ganguly, "End-to-end camera calibration for broadcast videos," in *Proceedings of the IEEE/CVF Conference on Computer Vision and Pattern Recognition*, 2020, pp. 13 627–13 636.

[12] M. Jaderberg, K. Simonyan, A. Zisserman *et al.*, "Spatial transformer networks," *Advances in neural information processing systems*, vol. 28, pp. 2017–2025, 2015.

[13] J. Chen and J. J. Little, "Sports camera calibration via synthetic data," in *Proceedings of the IEEE/CVF Conference on Computer Vision and Pattern Recognition Workshops*, 2019, pp. 0–0.

[14] I. Goodfellow, J. Pouget-Abadie, M. Mirza, B. Xu, D. Warde-Farley, S. Ozair, A. Courville, and Y. Bengio, "Generative adversarial networks," *Communications of the ACM*, vol. 63, no. 11, pp. 139–144, 2020.

[15] W. Jiang, J. C. G. Higuera, B. Angles, W. Sun, M. Javan, and K. M. Yi, "Optimizing through learned errors for accurate sports field registration," in *Proceedings of the IEEE/CVF Winter Conference on Applications of Computer Vision*, 2020, pp. 201–210.

[16] M. A. Fischler and R. C. Bolles, "Random sample consensus: a paradigm for model fitting with applications to image analysis and automated cartography," *Communications of the ACM*, vol. 24, no. 6, pp. 381–395, 1981.

[17] H. H. Bíscaro, S. M. Peres, and W. L. de Freitas, "Camera calibration for sport images: Using a modified ransac-based strategy and genetic algorithms," in *2013 IEEE Congress on Evolutionary Computation*. IEEE, 2013, pp. 1318–1325.

[18] I. De Falco, A. Della Cioppa, D. Maisto, and E. Tarantino, "Differential evolution as a viable tool for satellite image registration," *Applied Soft Computing*, vol. 8, no. 4, pp. 1453–1462, 2008.

[19] W. Ma, X. Fan, Y. Wu, and L. Jiao, "An orthogonal learning differential evolution algorithm for remote sensing image registration," *Mathematical Problems in Engineering*, vol. 2014, 2014.

[20] D. P. Kingma and J. Ba, "Adam: A method for stochastic optimization," *arXiv preprint arXiv:1412.6980*, 2014.

[21] G. Gallego and A. Yezzi, "A compact formula for the derivative of a 3-d rotation in exponential coordinates," *Journal of Mathematical Imaging and Vision*, vol. 51, no. 3, pp. 378–384, 2015.

[22] T.-Y. Lin, P. Goyal, R. Girshick, K. He, and P. Dollár, "Focal loss for dense object detection," in *Proceedings of the IEEE international conference on computer vision*, 2017, pp. 2980–2988.

[23] R. A. Sharma, B. Bhat, V. Gandhi, and C. Jawahar, "Automated top view registration of broadcast football videos," in *2018 IEEE Winter Conference on Applications of Computer Vision (WACV)*. IEEE, 2018, pp. 305–313.

[24] K. He, X. Zhang, S. Ren, and J. Sun, "Deep residual learning for image recognition," in *Proceedings of the IEEE conference on computer vision and pattern recognition*, 2016, pp. 770–778.